\documentclass{SCIS2019}

\usepackage{hhline}
\usepackage{xcolor}
\usepackage{colortbl}
\usepackage{microtype}
\usepackage{graphicx}
\usepackage{booktabs} 

\usepackage{hyperref}
\usepackage{mathtools}
\usepackage{bbm}
\usepackage[algo2e]{algorithm2e}
\usepackage[numbers]{natbib}
\usepackage{amsmath}

\usepackage{tikz}
\usepackage{pgfplots}
\usepackage{comment}
\usepackage{multirow}

\DeclarePairedDelimiter{\norm}{\lVert}{\rVert}

\def\balpha{\boldsymbol\alpha}

\def\bh{\mathbf{h}}

\def\br{\mathbf{r}}

\def\bu{\mathbf{u}}

\def\bw{\mathbf{w}}
\def\bx{\mathbf{x}}
\def\by{\mathbf{y}}
\def\bz{\mathbf{z}}
\def\bA{\mathbf{A}}

\def\bH{\mathbf{H}}

\def\bJ{\mathbf{J}}

\def\bM{\mathbf{M}}

\def\cL{\mathcal{L}}

\begin{document}
\ArticleType{RESEARCH PAPER}
\Year{2019}
\Month{}
\Vol{}
\No{}
\DOI{}
\ArtNo{}
\ReceiveDate{}
\ReviseDate{}
\AcceptDate{}
\OnlineDate{}


\title{Text Information Aggregation with Centrality Attention}{Text Information Aggregation with Centrality Attention}

\author{Jingjing Gong}{}
\author{Hang Yan}{}
\author{Yining Zheng}{}
\author{Qipeng Guo}{}
\author{Xipeng Qiu}{{xpqiu@fudan.edu.cn}}
\author{Xuanjing Huang}{}

\AuthorMark{Jingjing Gong}

\AuthorCitation{Jingjing Gong, Hang Yan, Yining Zheng, Qipeng Guo, Xipeng Qiu, Xuanjing Huang}


\address{
School of Computer Science, Fudan University, Shanghai 200433, China \\ 
Shanghai Key Laboratory of Intelligent Information Processing, Fudan University, Shanghai 200433, China\\ 
\\
\{jjgong, hyan11, ynzheng15, qpguo16, xpqiu, xjhuang\}@fudan.edu.cn
}

\abstract{A lot of natural language processing problems need to encode the text sequence as a fix-length vector, which usually involves aggregation process of combining the representations of all the words, such as pooling or self-attention.  However, these widely used aggregation approaches did not take higher-order relationship among the words into consideration. Hence we propose a new way of obtaining aggregation weights, called eigen-centrality self-attention. More specifically, we build a fully-connected graph for all the words in a sentence, then compute the eigen-centrality as the attention score of each word.
  The explicit modeling of relationships as a graph is able to capture some higher-order dependency among words, which helps us achieve better results in 5 text classification tasks and one SNLI task than baseline models such as pooling, self-attention and dynamic routing. Besides, in order to compute the dominant eigenvector of the graph, we adopt power method algorithm to get the eigen-centrality measure. Moreover, we also derive an iterative approach to get the gradient for the power method process to reduce both memory consumption and computation requirement.}


\maketitle

\section{Introduction}

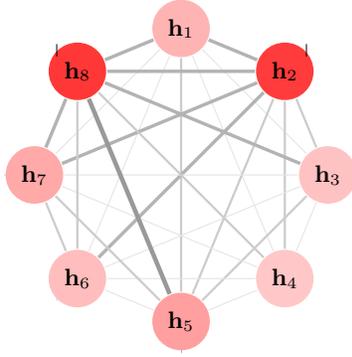
\begin{figure}
\centering
\begin{tikzpicture}[scale=0.65,font=\small\selectfont]

\draw[thick] (90: 3cm) node[circle,fill={rgb,255:red,255; green,180; blue,180}](1){$\mathbf{h}_1$};
\draw[thick] (45: 3cm) node[circle,fill={rgb,255:red,255; green,59; blue,59}](2){$\mathbf{h}_2$};
\draw[thick] (0: 3cm) node[circle,fill={rgb,255:red,255; green,194; blue,194}](3){$\mathbf{h}_3$};
\draw[thick] (-45: 3cm) node[circle,fill={rgb,255:red,255; green,199; blue,199}](4){$\mathbf{h}_4$};
\draw[thick] (-90: 3cm) node[circle,fill={rgb,255:red,255; green,159; blue,159}](5){$\mathbf{h}_5$};
\draw[thick] (-135: 3cm) node[circle,fill={rgb,255:red,255; green,189; blue,189}](6){$\mathbf{h}_6$};
\draw[thick] (-180: 3cm) node[circle,fill={rgb,255:red,255; green,169; blue,169}](7){$\mathbf{h}_7$};
\draw[thick] (-225: 3cm) node[circle,fill={rgb,255:red,255; green,55; blue,55}](8){$\mathbf{h}_8$};

\draw[line width=0.30mm,draw=black!20](1)--(1);
\draw[line width=0.45mm,draw=black!30](1)--(2);
\draw[line width=0.15mm,draw=black!10](1)--(3);
\draw[line width=0.15mm,draw=black!10](1)--(4);
\draw[line width=0.30mm,draw=black!20](1)--(5);
\draw[line width=0.15mm,draw=black!10](1)--(6);
\draw[line width=0.15mm,draw=black!10](1)--(7);
\draw[line width=0.45mm,draw=black!30](1)--(8);
\draw[line width=1.65mm,draw=black!110](2)--(2);
\draw[line width=0.30mm,draw=black!20](2)--(3);
\draw[line width=0.30mm,draw=black!20](2)--(4);
\draw[line width=0.30mm,draw=black!20](2)--(5);
\draw[line width=0.45mm,draw=black!30](2)--(6);
\draw[line width=0.45mm,draw=black!30](2)--(7);
\draw[line width=0.45mm,draw=black!30](2)--(8);
\draw[line width=0.15mm,draw=black!10](3)--(3);
\draw[line width=0.15mm,draw=black!10](3)--(4);
\draw[line width=0.30mm,draw=black!20](3)--(5);
\draw[line width=0.15mm,draw=black!10](3)--(6);
\draw[line width=0.15mm,draw=black!10](3)--(7);
\draw[line width=0.45mm,draw=black!30](3)--(8);
\draw[line width=0.15mm,draw=black!10](4)--(4);
\draw[line width=0.15mm,draw=black!10](4)--(5);
\draw[line width=0.15mm,draw=black!10](4)--(6);
\draw[line width=0.15mm,draw=black!10](4)--(7);
\draw[line width=0.30mm,draw=black!20](4)--(8);
\draw[line width=0.45mm,draw=black!30](5)--(5);
\draw[line width=0.15mm,draw=black!10](5)--(6);
\draw[line width=0.30mm,draw=black!20](5)--(7);
\draw[line width=0.60mm,draw=black!40](5)--(8);
\draw[line width=0.30mm,draw=black!20](6)--(6);
\draw[line width=0.30mm,draw=black!20](6)--(7);
\draw[line width=0.30mm,draw=black!20](6)--(8);
\draw[line width=0.30mm,draw=black!20](7)--(7);
\draw[line width=0.45mm,draw=black!30](7)--(8);
\draw[line width=1.65mm,draw=black!110](8)--(8);
\end{tikzpicture}
\caption{Eigenvector centrality of a fully-connected word graph. Darker color of a node indicates it is more important, the thickness of the edge represents the connected strength. The centrality is computed based on the connected relationship among all the words.}
\label{fig:eigenvector-centrality}
\end{figure}

\label{sec:intro}
  In the field of natural language processing (NLP), aggregating word representation vectors of variable lengths into one or several vectors is crucial to the success of various NLP tasks. This aggregation process prevails in text classification tasks \cite{tang2015learning,socher2013recursive} and encode-based SNLI tasks \cite{Bowman2015snli}.

  There are a bunch of work try to represent text sequence as a fixed-size vector. Various methods differentiate from each other mostly in the way they encode the text sequence and aggregate word representations. In this paper, we mainly focus on  aggregation operation, namely how combining the a sequence of vectors. \cite{socher2013recursive} recursively encodes a sentence into a fixed size vector, the process can be viewed as concurrent encoding and aggregating. \cite{pfliu15MultiTimeScale} treat the hidden states of the last time step in a variant of LSTM as the encoded vector of a sentence. \cite{kim2014convolutional} combines convolutional neural networks and max-pooling to solve text classification tasks. For the purpose of alleviating the remembering burden on encoding layers, \cite{zhanlin17structuredSelf} design an attention mechanism to calculate a weight for each encoded states. The aforementioned aggregation methods lack of integrating global information into the decision of the weights, \cite{jjgong18infaggCapsule} enhance the representational ability of aggregated vector by dynamic routing, it is iteratively refining the routing weights taking more context information into consideration after each iteration.

  Among several aggregation methods, self-attention mechanism \cite{zhanlin17structuredSelf,yang2016hierarchical} shows its advantage of flexibility and effectiveness.
  However, the self-attention assigns the weight to each word according to its own representation and a query vector, and ignores the impact of the contextual words. Intuitively, the importance of a word in a sentence should depend on the relationships among all the words.



  In this paper, we propose a new method to calculate the attention weight by explicitly modeling an intensive relationship between words in the sequence. We first construct a fully-connected word graph in which the edges are the relationships between words, and vertices are words in the sequence. We then apply centrality algorithm on this graph and then the importance score of each vectors will emerge.
  The ``centrality'' is the importance of each node in a graph. The assumption is that each node's centrality is the sum of the centrality values of the nodes that it is connected to.
  Figure \ref{fig:eigenvector-centrality} gives an intuitive illustration of the centrality of a fully-connected graph.

  The contribution of this paper can be summarized as follows:

  \begin{enumerate}
    \item We have identified that most aggregation methods are independent of contextual words, thus propose a more context-aware approach to obtain attention weights, which considers an intense interaction between words.
    \item We utilize power method to get the dominant eigenvector of the graph, but constructing computation graph during the iterative process of power method makes the back propagation very low-efficient. Therefore, we derive an approach to approximate the gradient of the power method which will significantly reduce memory consumption and the back-propagation computation complexity.
    \item We justify the use of ``stop gradient before convergence" trick to reduce the memory utility while still utilizing auto-differentiation frameworks.
  \end{enumerate}



  \section{Background}
  \subsection{Problem Definition}
      Commonly, the sentence encoding process aims to find a fix-length vector to represent the meaning of a variable-length sentence, which consists of the following three layers.

      \paragraph{Embedding Layer}
      Given a sentence $S = x_1, x_2,...,x_n$, most sentence encoding models first transform each word in $S$ into $d$ dimensional dense vectors. Thus, the sentence is represented by a matrix into  $X = [\bx_1, \bx_2,...,\bx_n]\in \mathbb{R}^{d\times n}$. Some pre-trained embedding, such as word2vec \cite{mikolov2013word2vec} or GloVe \cite{pennington2014glove}, can be used for a good initialization.
      However, these pre-trained embeddings lack the awareness of words' context, ELMO \cite{peters2018Elmo} solves this problem by generating word embeddings by the language model. To have fair comparison with previous work, we use static embedding Glove in this paper.

      \paragraph{Fusion Layer}
        Fusion Layer is responsible for modeling the semantic compositions between words, RNN, CNN and Transformer are universally adopted. No matter what kind of structures are applied, the basic form of fusion layer can be formalized as \\
        \begin{align}
          \bh_1, \bh_2,..., \bh_n &= \text{fuse}(\bx_1,\bx_2, ..., \bx_n),\\
          \bH &= [\bh_1, \bh_2,..., \bh_n]
        \end{align}
        During fusion phase, $x_t$ as well as all other word representations should have an impact on the value of $h_t$. In this paper, BiLSTM, which not only considers the forward contextual correlation but also accounts for information flow from the reverse direction, serves as the Encoder Layer. We concatenate the forward hidden state and the backward hidden state in the last layer of the BiLSTM.

      \paragraph{Aggregation Layer}

        Different sentences vary in length, the aggregation of variable-length representations $H$ into a fixed-length vector is necessary in most NLP tasks.
        A general aggregation method is to assign a weight $\alpha_i$ to each word, then the summarized vector $\bar{\bh} \in \mathbbm{R}^d$ is
    \begin{align}
              	\bar{\bh} &=\sum_{i=1}^n {\alpha_i \bh_i},\label{eq:weighted_sum}
    \end{align}
    where the weight $\alpha_i$ can be determined statically or dynamically according to different strategies.

     Following are four commonly used aggregation layers.


        {\textit{Max-Pooling}} takes the maximum value of encoded representations in each dimension, it can be expressed as
        \begin{align}
          \bar{\bh}&=\max([\bh_1,\bh_2,\cdots,\bh_n])
        \end{align}

        {\textit{Average Pooling}} set the weight of each word to $\alpha_i=\frac{1}{n}$.  The average pooling is an aggregation method that feature vectors are weakly involved in determining the process of aggregation. Each word can only influence how it is aggregated by its quantity, because the weight of each word is inverse proportional to its length.

        {\textit{Self-Attention}} dynamically assigns a weight to each word according its representation \cite{zhanlin17structuredSelf,yang2016hierarchical} , then sums up all the words according to their weights. A trainable query $\mathbf{q}$ of the same dimensions as the encoded representation is used to calculate this weight. Followings are self-attention's formula
        \begin{align}
            \alpha_i&=\frac{\exp(\mathbf{q}^T\bh_i)}{\sum_{j=1}^n{\exp(\mathbf{q}^T\bh_j)}}\label{eq:softmax}
        \end{align}

        Though being more complex than average pooling, self-attention the calculation of the score also neglects other words in the sentence, and the only interactions among the words is the softmax normalization process.

        \textit{Dynamic Routing}
        is a mechanism that transfers the information of $H$ into a certain number of vectors $V$. Normally there are two manners to further exploit vectors $V$, one is to treat each vectors as the representation for each classes \cite{sabour2017dynamic}, the other is to concatenate vectors in $V$ into a fixed-length vector \cite{zhang2018multi}. The detail dynamic routing process can be found in \cite{sabour2017dynamic}.

    \subsection{Eigenvector Centrality} \label{sec:Eigenvector-Centrality}

    The centrality measure is a kind of quantitative symbol of influence(importance) value of a node in a network. There exist several popular centralities: (1) In-degree centrality, a node's in-degree centrality is decided by the number of nodes pointing to it. (2) Closeness centrality, to get a node's closeness centrality, first sum up distances between the node and its connected nodes, then compute the sum's reciprocal. (3). Betweenness centrality, the number of times a node appears in the shortest path between other two nodes. (4) Eigenvector centrality, which we will describe in later section.
     \cite{fletcher2018structure,bonacich1972factoring,bonacich2007some,lohmann2010eigenvector} utilize eigenvector centrality to capture intrinsic neural architectures through fMRI data. Eigenvector centrality is regarded as an indicator of global prominence of a feature in \cite{roffo2016features}.

    In this work, we believe the eigen-centrality of a word in a sentence can indicate the importance of this word, and we take its eigen-centrality value as its weight to aggregate the sentence.

  \section{Eigen-Centrality Self-Attention}
    Given a sentence with its representation $H=[\bh_1, \bh_2, \cdots, \bh_n] \in \mathbb{R}^{d \times n}$, the aggregation is to assign a weight $\alpha_i$ to each word.
    Intuitively, the weight of each word should be determined by the relations between all the words in the sentence.

    Therefore, we construct an adjacency matrix $\bA \in \mathbb{R}^{n \times n}$ to describe the connectivity between words in the vector list. Each element in $\bA \in \mathbb{R}^{n \times n}$ are formulated as follows:
    \begin{align}
      \label{eq:adaj}
      A_{ij} &= f(\bh_i, \bh_j),
    \end{align}
    where $A_{ij}$ is the $i$th row $j$th column of $\bA$, function $f$ is a scalar connectivity function that its result is strictly positive. Connectivity function $f$ should be designed to be trainable functions that are able to capture the relationship between a pair of components and describe it as connectivity intensity. In this paper, the $f$ is chosen to be a two layer fully connected neural network followed with a column-wise softmax normalization to make the adjacency matrix $\bA$ a left stochastic matrix.

    \subsection{Calculating Attention Score via Eigen-Centrality}

    Given the adjacency matrix $\bA$ of a sequence of words, where $\bA_{ij}$ denotes a connection strength from word $i$ and word $j$.
    A larger value of $\bA_{ij}$ indicates the stronger relation between word $i$ and word $j$.

    The importance of a word can be measured by centrality. Here, we use the eigenvector centrality to calculate the importance of a word with respect to other words.  A high score of a word suggests that there are large connected weights between it and the other important nodes.
    A well-known application of eigenvector centrality is Google's PageRank algorithm.

    The relative centrality score of $i$th word can be defined as:
      \begin{align}
        \label{eq:cent-conn}
        \alpha_i &= \frac{1}{\lambda}\sum_{j}{A_{i, j}\alpha_j}
      \end{align}
      where $\alpha_i$ denotes the relative importance measure of $i$th vertex, $\lambda$ is a constant, note that both $\alpha_i$ and $\lambda$ are unknowns that need to be solved. Let $\balpha = (\alpha_1, \alpha_2, \cdots, \alpha_n)^\top$, Eq. \ref{eq:cent-conn} can be formulated as vector notation:
      \begin{align}
        \label{eq:cent_alg}
        \bA\balpha &= \lambda\balpha
      \end{align}
      where $\balpha$ happens to be an eigenvector of adjacency matrix $\bA$, note that only the eigenvector with all it's values positive will satisfy the requirement. And according to Perron-Frobenius Theorem \cite{frobenius1912matrizen, perron1907theorie}, when all entries $\bA_{ij}$ in the adjacency matrix $\bA$ is strictly positive, then there exists and only exists one eigenvector $\balpha = (\alpha_1,...,\alpha_n)^\top$ of $\bA$ with a dominant eigenvalue $\lambda$ such that all components of $\balpha$ are positive: $\bA\balpha = \lambda\balpha, \forall i \quad \alpha_i > 0$.
      It is known in the literature under many variations, such as the Perron vector, Perron eigenvector, Perron-Frobenius eigenvector, leading eigenvector, or dominant eigenvector.

    \subsection{Obtaining Dominant Eigenvector via Power Method}

    Now we need to solve the equation in Eq. \ref{eq:cent_alg} to get the dominant vector $\balpha = (\alpha_1,...,\alpha_n)^\top$. Since we only need the dominant eigenvector, we apply the Power Method algorithm to obtain the dominant eigenvector of the adjacency matrix with moderate computational cost.

    \begin{algorithm}[t]
      \SetKwInOut{Input}{Input}\SetKwInOut{Output}{OutPut}
      \Input{Connectivity matrix $\bA \in \mathbbm{R}^{n \times n}$, \\
              initialize random vector $\by = \bz \in \mathbbm{R}^n$;}

      \Repeat{$\norm{\by - \theta\balpha}_2 \leq \epsilon \abs{\theta}$}{
        $\balpha = \frac{\by}{\norm{\by}_2}$\;

        $\by = \bA\balpha$\;

        $\theta = \balpha^\top\by$\;
      }
      \tcp{Stop when converge}
      \Output{Dominant eigenvalue: $\lambda = \theta$ \\
              and corresponding eigenvector: $\balpha$}
      \caption{Power Method}\label{alg:power-method}
    \end{algorithm}
    \cite{lin2010power} have made convergence analysis on power method convergence and have shown that power method good at approximating eigenvalue and eigenvectors.
    In the scope of this paper, the adjacency matrix is strictly positive. The Perron-root or dominant eigenvalue have algebraic and geometric multiplicities both be one. Unless the initial vector $\bz$ is strictly perpendicular with respect to the dominant eigenvector, a randomly initialized start vector could converge to the dominant eigenvector. Thus we initialize the random $\bz$ with all positive components to prevent a perpendicular initialization. We will discuss the initialization of this initial vector in detail in Appendix \ref{ap:zero-grad} \ref{ap:analy2}.

    \subsection{Gradient Computation for Power Method}
      With auto-differentiation tools we can easily compute the gradient for a normal operation. But the power method need to run for ideally infinite iterations to get an accurate approximation of gradient, to use auto-differentiation we need to store intermediate states for every iteration step, and the backward is also to be compute as many steps as forward iteration.
      In an ideal case where we run power method for large number of steps, the memory consumption would be unbearable, and the backward process would also double the computation as needed in forward iteration.

      In this paper we prove theorem \ref{le:zero-grad} that when the power method runs for ideally infinite steps, initial vector $\bz$ receive neglectable amount of gradient which means the initial $\bz$ won't affect the result of gradient, this also indicate that  gradient received in early steps is neglectable. which further indicate that only last few steps of iteration matters. this suggests that when convergence criteria have reached (we call it convergence phase), we can run extra steps to obtain a good approximation of the gradient (we call it gradient iteration phase), also since states in early step have little influence on the final gradient, in convergence iteration phase we can iterate without storing any intermediate steps.

      \begin{theorem}
        \label{le:zero-grad}
        Given an adjacency matrix $\bA$ with its component $\quad A_{ij} > 0 \quad\forall i,j$, randomly initialize $\bz$ with positive entries as the initial vector of power method for infinite steps. The partial derivative of the output eigenvector $\balpha$ with respect to $\bz$ (or early steps of $\balpha$) is $\mathbf{0}$.
      \end{theorem}

      Since we can approximate gradient with last few steps of iteration, we know that the eigenvector would have been long converged, which means we even don't need to store the intermediate state in gradient phrase iterations, since every intermediate eigenvector would be approximately the same, and the adjacency matrix $\bA$ is also constant.

      Thus, we are also introducing a reverse iteration approach (Eq. \ref{eq:iter-grad}) to calculate the gradient of power method, in which if the approximation of the dominant eigenvector is good enough, we can bound the error of the approximated gradient to a small margin. Moreover, none intermediate states will be kept, since after infinite steps the intermediate state would be the same since it would have long be converged to the dominant eigenvector. We will prove Theorem \ref{th:grad} in Appendix \ref{ap:grad}.

      \begin{theorem}
        \label{th:grad}
        Given an adjacency matrix $\bA$ with all it's components $\quad A_{ij} > 0 \quad\forall i,j$, it's corresponding dominant eigenvalue $\lambda$ and corresponding eigenvector $\balpha$ computed with power method algorithm for infinite steps. The derivative of loss $\cL$ with respect to $\bA$ can be computed as:

        \begin{align}\label{eq:iter-grad}
          \frac{\partial \cL}{\partial \bA} = \lim_{N \to \infty} \sum_{k=0}^{N}{{\boldsymbol\gamma}^{\top} (\bJ_{\balpha})^k \bJ_{\bA}}
        \end{align}

        where, $\boldsymbol\gamma$ is the partial derivative of $\cL$ with respect to last step of $\balpha$, $\bJ_{\balpha}$ is the jacobian matrix of $\balpha$ with respect to previous step of $\balpha$, $\bJ_{\bA}$ is the jacobian matrix of $\balpha$ with respect to previous step of $\bA$.($\bJ_{\balpha}$ and $\bJ_{\bA}$ depend only on $\bA$, $\balpha$ and $\lambda$.)

      \end{theorem}

      \section{Application to NLP Tasks}
      \begin{figure*}[t!]
        \centering
        \subfloat[Flatten Model for Sentence Classification]{
        \includegraphics[width=0.40\textwidth]{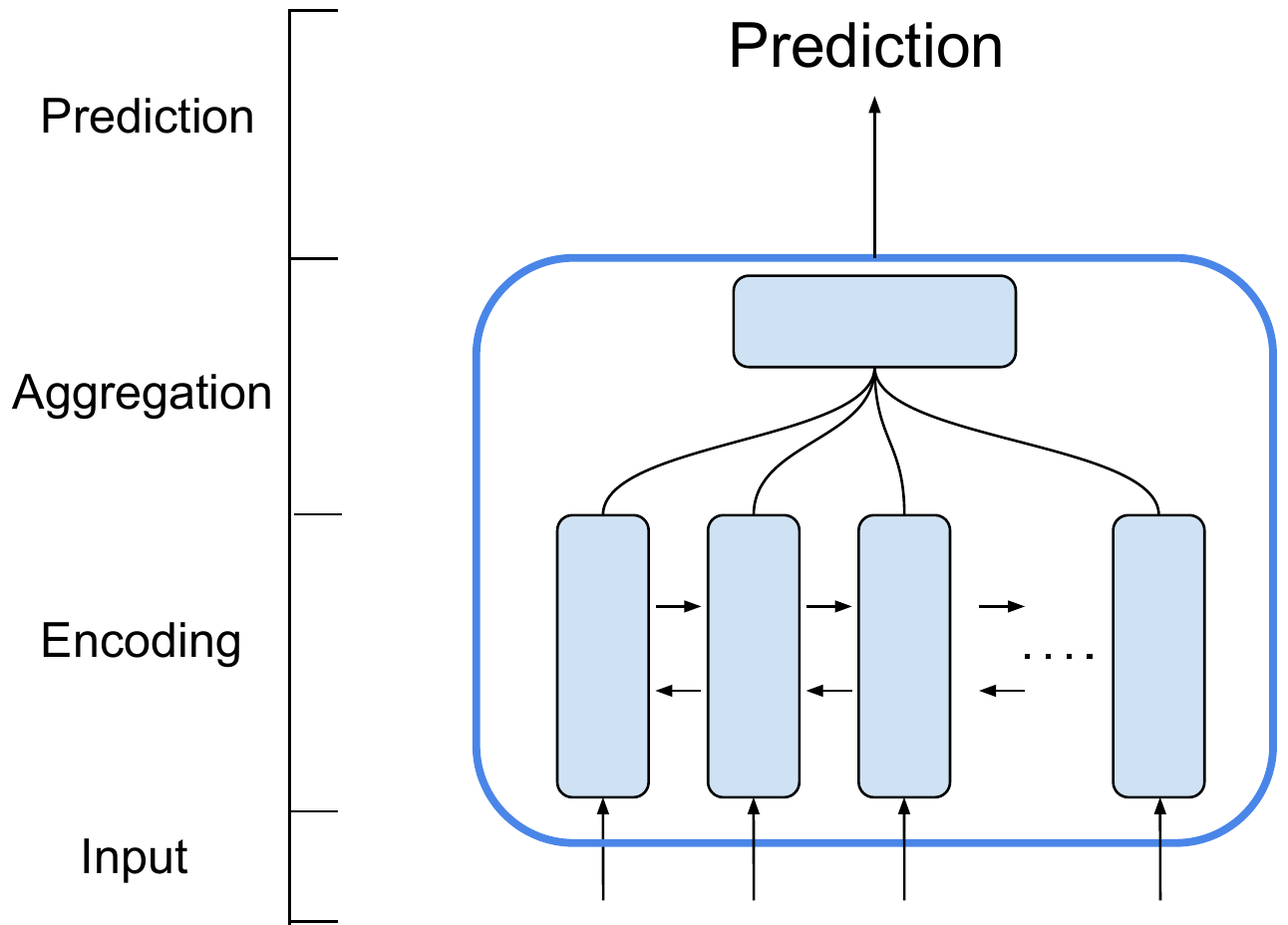} \label{fig:flatten}
        }
        \hspace{3em}
        \subfloat[Hierarchical Model for Document Classification]{
        \includegraphics[width=0.47\textwidth]{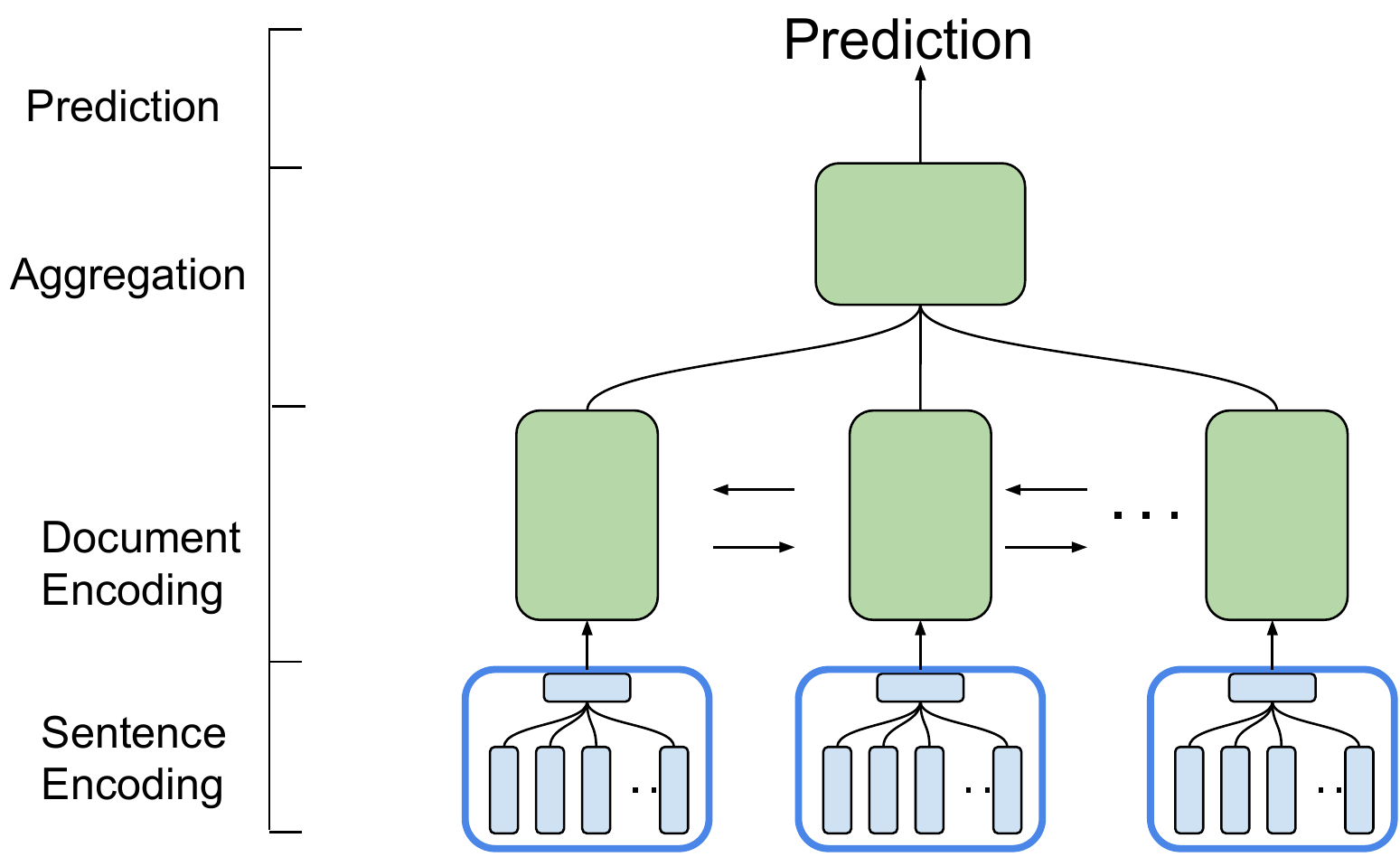} \label{fig:hierarchical}
        }
        \caption{Text classification architectures. Where left is the architecture for SST-1 and SST-2 datasets, right figure is the architecture for IMDB, Yelp datasets. they share the same sequence encoding model which is BiLSTMs and the aggregation method --- eigen-centrality attention.}\label{fig:models-classification}
      \end{figure*}

      \begin{figure}[t!]
        \centering
        \includegraphics[width=0.45\textwidth]{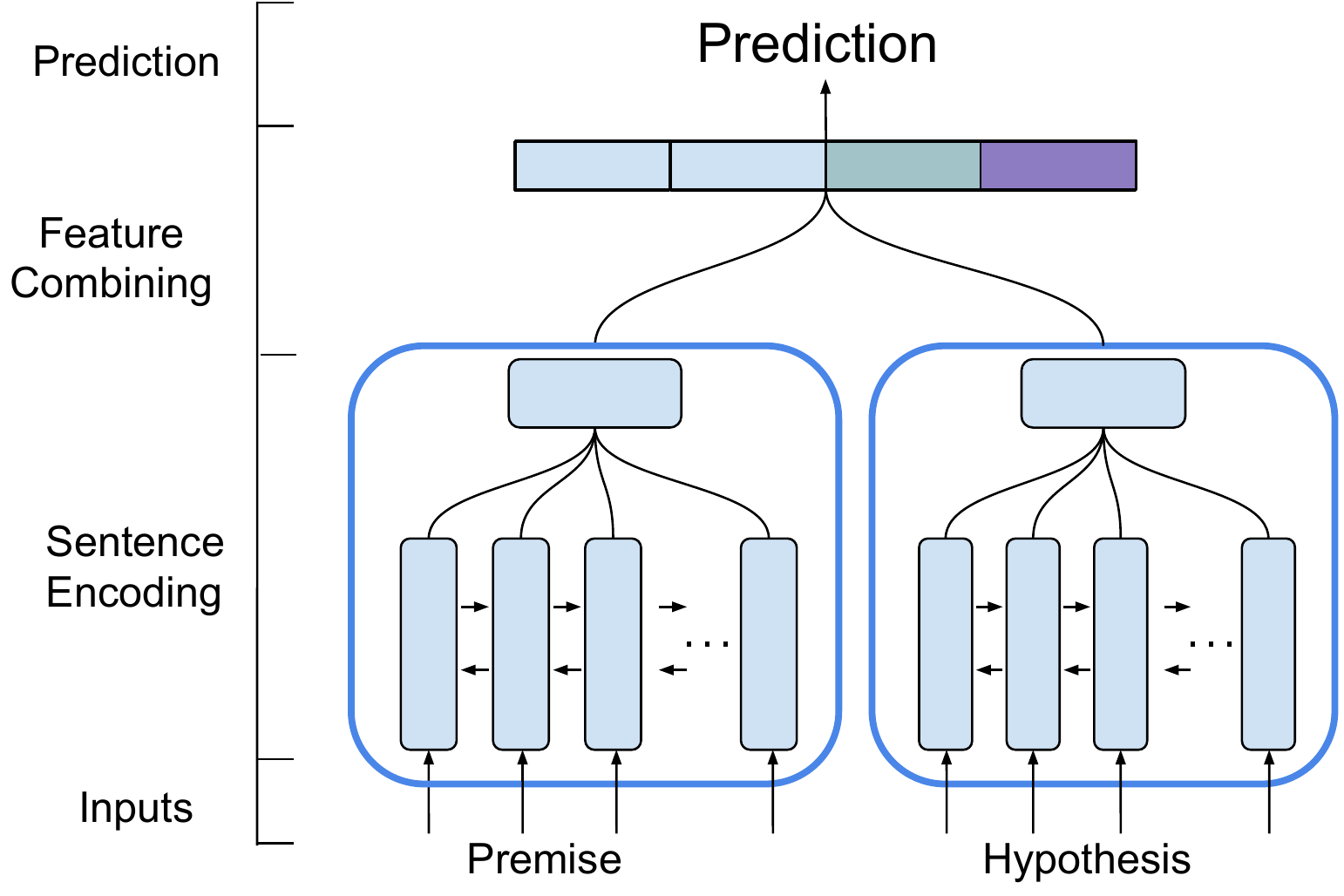}
        \caption{NLI Architecture, the encoder is the same as in Fig. \ref{fig:flatten}.}\label{fig:nli-model}
      \end{figure}

      There are three kinds of tasks in this paper, sentence level classification, document level classification and encoder based texture entailment task. The sequence encoding is exactly the same but with a different network structure.
      \paragraph{Flatten Sentence Classification Architecture}
        In sentence level classification, we adopt a flatten structure which encodes the word embedding sequence with a BiLSTMs followed with an aggregation operation. The aggregation operation in this paper is the proposed eigen-centrality attention. The architecture schematic is shown in Fig. \ref{fig:flatten}.
      \paragraph{Hierarchical Document Classification Architecture}
        In document level classification we take a hierarchical approach \cite{yang2016hierarchical}, we first encode sentences and aggregate them as sentence embeddings with BiLSTMs followed with the proposed eigen-centrality attention, and then encode the sentence representations as document representations with BiLSTMs followed with an eigen-centrality attention. The hierarchical classification architecture is shown in Fig. \ref{fig:hierarchical}.

      \paragraph{NLI Architecture}
        In texture entailment task, we first separately encode the premise and hypothesis sentences and get the sentence representations and then combine the two representations as one to represent the relationship between the two sentences. The feature combining function is defined as:
        \begin{align}
          \br = [\br_p; \br_h; \abs{\br_p - \br_h}; \br_p * \br_h]
        \end{align}
        where $\br$ is the final representation of the relation between premise sentence and hypothesis sentence, $\br_p$ is the premise sentence representation and $\br_h$ is the hypothesis sentence representation.
        The schematic of NLI architecture is shown in Fig. \ref{fig:nli-model}.

      Having obtained the representations, we feed the representation vector through a feed forward neural network followed with a softmax normalization function to get the probability vector. Then we optimized a following cross-entropy objective function:
      \begin{equation}
        \mathcal{J}(\Theta) = \frac{1}{N}\sum_{i=1}^{N}\log p(Y_i;\Theta),
      \end{equation}
      where $\Theta$ is trainable parameters in the model, $Y_i$ is the target label of $i$th example in the batch.

\begin{table*}[t!] \setlength{\tabcolsep}{3pt}
      \centering
      \begin{tabular}{lccccccc}
        \toprule
      \textbf{Dataset} &\textbf{Type} &\textbf{Train Size} & \textbf{Dev. Size} &\textbf{Test size} & \textbf{Classes}  &\textbf{Averaged Length}   &\textbf{Vocabulary Size}\\
        \midrule 
        Yelp 2013         &Document      &62522       &7773    &8671     &5    &189 	&29.3k\\
        Yelp 2014         &Document      &183019      &22745   &25399 	 &5    &197     &49.6k\\
        IMDB          	  &Document      &67426      &8381   &9112 	     &10    &395    &61.1k\\
        SST-1         	  &Sentence      &8544      &1101   &2201	     &5    &18      &16.3k\\
        SST-2         	  &Sentence      &6920      &872   &1821	     &2    &19      &14.8k\\
        SNLI              &NLI-task      &549k      &9.8k    &9.8k      &3      &11     &36k \\
        \bottomrule
      \end{tabular}
      \caption{Statistics of the five classification datasets used in this paper}
      \label{tab:dataset-stat}
    \end{table*}

  \section{Experiments}

    \begin{table*}[h] \setlength{\tabcolsep}{3pt}
          \centering
          \begin{tabular}{lcccccc}
            \toprule
            &\textbf{Yelp-2013} &\textbf{Yelp-2014} & \textbf{IMDB} &\textbf{SST-1} & \textbf{SST-2} &\textbf{SNLI}\\
            \midrule 
            Embedding size           		&300    	&300      &300    &300     &300   &300    \\
            LSTM hidden unit         		&300    	&300      &300    &300     &300   &300    \\
            connectivity hidden units   &50       &50       &50     &50      &50    &30     \\
            Regularization rate        	&1e-6     &1e-6     &1e-6   &1e-6    &1e-6  &1e-20  \\
            Initial learning rate       &0.0001   &0.0001   &0.0001 &0.0003  &0.0003&0.0001 \\
            learning rate decay        	&0.9      &0.9      &0.9   	&0.95    &0.95  &0.95   \\
            learning rate decay steps   &2000     &5000     &1000   &500     &500   &20000  \\
            Initial Batch size         	&64      	&64     	&32   	&128     &128   &128    \\
            Batch size low bound        &32      	&16     	&32   	&32      &16    &128    \\
            Dropout rate         			  &0.6      &0.6      &0.4   	&0.6     &0.6   &0.2    \\

            \bottomrule
          \end{tabular}
          \caption{Detailed hyper-parameter settings}
          \label{tab:hypersetting}
        \end{table*}

        \begin{figure*}[t!]
        \centering
        \includegraphics[width=0.90\textwidth]{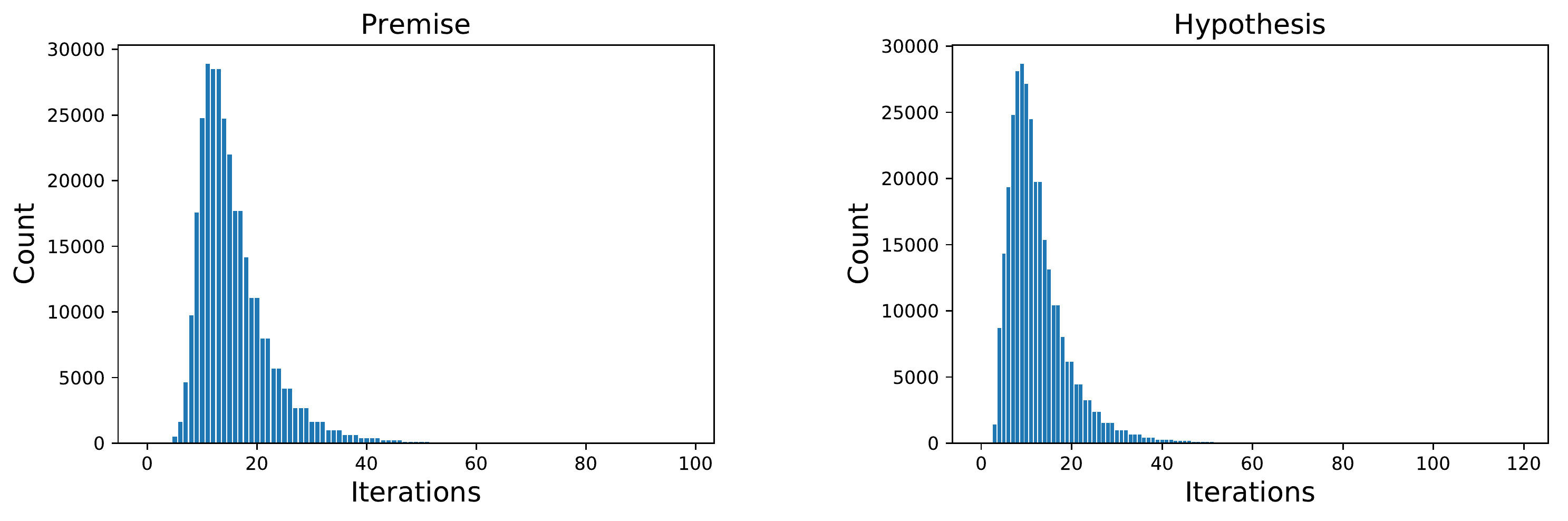}
        \caption{Power method convergence statistics, it shows that for majority samples the power method converge before $200$ step. Note that the vertical axis which denotes the sample count and the horizontal axis denotes the iteration steps needed to reach convergence.}\label{fig:converge-stats}
      \end{figure*}

    \begin{table*}[t] 
      \centering\small
      \begin{tabular}{lccccc}
        \toprule
        &\textbf{Yelp-2013} &\textbf{Yelp-2014} &\textbf{IMDB} & \textbf{SST-1}  &\textbf{SST-2}  \\
            \midrule
        RNTN+Recurrent~\cite{socher2013recursive} &57.4	&58.2	&40.0	&-	&-	\\
        CNN-non-static~\cite{kim2014convolutional}&-		&-		&- 		&48.0	&87.2	 \\
        Paragraph-Vec~\cite{le2014distributed}  	& -		& - 	& - 	& 48.7 & 87.8  \\
        MT-LSTM (F2S)~\cite{pfliu15MultiTimeScale} &- &- 	&- 		&49.1 	&87.2  \\
        UPNN(np UP)~\cite{tang2015learning} 		&57.7	&58.5	&40.5	&- 	&-  \\
        UPNN(full)~\cite{tang2015learning} 		& 59.6	&60.8 	& 43.5 &-  &-   \\
        Cached LSTM~\cite{jcxu16cachedLSTM}  &59.4  &59.2  &42.1  &- &-  \\
        Standard DR-AGG~\cite{jjgong18infaggCapsule}   &62.1   &63.0   &45.1   &50.5   &87.6 \\
        Reverse DR-AGG~\cite{jjgong18infaggCapsule}    &61.6   &62.5   &44.5   &49.3   &87.2 \\
        \midrule
        Max pooling       &61.1   &61.2   &41.1   &48.0   &87.0      \\
        Average pooling   &60.7   &60.6   &39.1   &46.2   &85.2      \\
        Self-attention    &61.0   &61.5   &43.3   &48.2   &86.4		 \\

        \midrule
          This work    &\textbf{63.7}   &\textbf{64.2}   &\textbf{48.2}   &\textbf{51.6}   &\textbf{88.5}	 \\
        \bottomrule
      \end{tabular}
      \caption{Experimental results comparison on five classification datasets.}
      \label{tab:cls-results}
    \end{table*}

    We have empirically conducted experiments on 5 text classification tasks and a Natural Language Inference task (SNLI). Detailed dataset statistics is shown in table \ref{tab:dataset-stat}.
    \subsection{Datasets}

      \textbf{SST-1}
        Stanford Sentiment Treebank is a movie review dataset which has been parsed and further splited to train/dev/test set \cite{socher2013recursive}. For each example in the dataset, there exists only one sentence and a label associated with it. And the labels can be one of \{negative, somewhat negative, neutral, somewhat positive, positive\}.

      \textbf{SST-2}
        This dataset is a binary-class version of SST-1, with neutral reviews removed and the remaining reviews categorized to either negative or positive.

      \textbf{IMDB}
        is a movie review dataset extracted from IMDB website. It is a multi-sentence dataset that for each example there are several review sentences. A rating score range from 1 to 10 is also associated with each example.

      \textbf{Yelp reviews} Yelp-2013 and Yelp-2014 are reviews from Yelp, each example consists of several review sentences and a rating score range from 1 to 5 (higher is better). Note that we use the same document level datasets as provided in \cite{tang2015learning}

      \textbf{SNLI} The SNLI dataset is a collection of 570k human-written sentence pairs which are called premise and hypothesis sentence, for each sentence pair there exists a relationship labeled as entailment, contradiction and neutral, and the objective of this task is to predict the relationship between premise and hypothesis sentence. Note that in this paper we encode sentences without interaction.

    \subsection{Implementation Details}

       We adopt adam optimizer\cite{kingma2014adam} to optimize all trainable parameters. To prevent over fitting, we also adopt regularization tricks such as weight decay, dropout are applied on the word embeddings and in the hidden layers of the final feed forward neural net.

      The hidden units of the connectivity function $f$ in Eq. \ref{eq:adaj} is set to $30$ for SNLI dataset, for classification tasks it's set to $50$. The word embedding dimensions are unanimously set to $300$, the hidden units of the BiLSTMs are set to $(300+300)$
      We set the stop factor $\epsilon = 1e^{-10}$ and set convergence phase max power iteration to $200$ steps, as shown in Figure \ref{fig:converge-stats}, maximum $200$ steps of iteration is more than enough for most of the samples. Note that in convergence phase we don't store any intermediate states, and no gradient is required, after convergence we run extra $20$ steps to approximate the gradient.
      Detailed hyper-parameter setting is given in the Table \ref{tab:hypersetting}.

     \begin{table}[t!] 
    \centering
    \begin{tabular}{lccccc}
      \toprule
      \textbf{Model} &\textbf{Test Accuracy}  \\
      \midrule
        %
        Max pooling~\cite{Conneau17sentrep}              &84.5     \\
        Intra-attention~\cite{yLin16blstm}               &84.2     \\
        Self-attention~\cite{zhanlin17structuredSelf}    &84.4     \\
      \midrule
        Average pooling                                     &83.4      \\
      \midrule
        This work    &\textbf{85.3} \\
      \bottomrule
    \end{tabular}
    \caption{Experimental results comparison on SNLI.}
    \label{tab:snli-results}
  \end{table}
    \subsection{Experimental Results}
      \label{exp}
      We evaluate our proposed aggregation method on five text classification datasets and a Natural Language Inference dataset (SNLI). In which IMDB, Yelp-2013 and Yelp-2014 are document level datasets, SST-1, SST-2 are sentence level datasets, SNLI is the Natural Language Inference dataset. since average pooling, self-attention, max-pooling are most related to our work, we mainly compare our work to those baselines, and compare our result on SNLI to encoder based models.

      The detailed comparison on classification datasets is shown in Table \ref{tab:cls-results}, it shows that our proposed aggregation have outperforms all methods in comparison. and specifically a considerable improvement have achieved on IMDB dataset, 3.7\% absolute accuracy improvement compared to DR-AGG.

      Detailed result comparison on SNLI is shown in Table \ref{tab:snli-results}, it says that our proposed eigen-centrality attention is effective and out-performed max-pooling, mean-pooling and self-attention.

    \subsection{Visualization of latent graph}
      \label{sec:visual}
      \begin{figure*}[t]
        \centering
        \subfloat[Premise]{
        \includegraphics[width=0.90\textwidth]{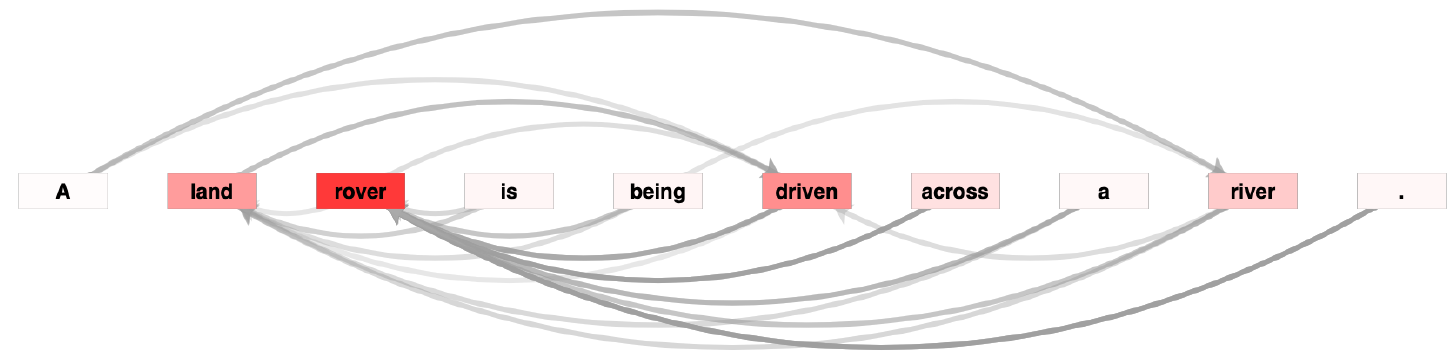} \label{fig:visual_p}
        }\\
        \subfloat[Hypothesis]{
        \includegraphics[width=0.65\textwidth]{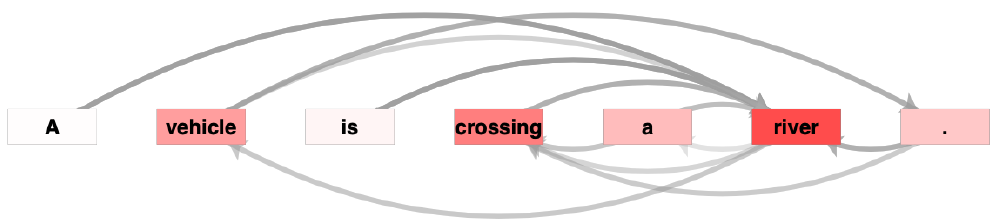} \label{fig:visual_h}
        }
        \caption{Latent Graph Visualization of a sample in SNLI dataset. The top sentence \ref{fig:visual_p} is the premise sentence and the bottom \ref{fig:visual_h} one is the hypothesis sentence. The color density of words indicate the importance of each word in this sentence, and the edge is the connection between words, a darker color indicate a more intense connection.}\label{fig:latent-graph}
      \end{figure*}

       One important feature in our work is that, we have an intermediate graph, and from this graph we can apply power method to get the eigen-centrality measure of each node in this graph. Shown in Fig. \ref{fig:latent-graph} is a sample from SNLI dataset where top sentence \ref{fig:visual_p} is the premise sentence and the bottom one \ref{fig:visual_h} is the hypothesis sentence. The color density of words indicate the importance of each word in this sentence, and the edge is the connection between words, a darker color indicate a more intense connection. Note that we first have the connection then derives the importance measure, not the other way around. In the example the word ``rover" has the most attention, the attention is attributed to the nodes which are connected to it, if the node that connects to it is important and has intense connection, then the node contributes more to the word. This distinguishes eigen-centrality attention from almost all existing aggregation methods, which have identified as weak contextual dependency as described in section \ref{sec:intro}.

 \section{Discussion}
   \subsection{Connection to Self-Attention}
   The self-attention is actually a subspace inside the space of eigen-centrality attention. Eq. \ref{eq:adaj} could be optimized to a subspace so that it only cares about $\bh_i$, then:
   \begin{align}
     A_{ij} &= g(\bh_i)
   \end{align}

   If the function $f$ is collapsed into $g$ and $g$ being a linear function as in self-attention, with a column softmax normalization, the eigenvector is exactly the same as the assigned weights in self-attention.

   In self-attention the score of a component depend on the component itself (if not consider the final normalization), while in eigen-centrality attention, the score of a component have a recursive dependency on all components connected to it.

   \subsection{Connection to Dynamic Routing}
      Dynamic routing \cite{jjgong18infaggCapsule} and eigen-centrality attentions share the same attribute that, which source component is to pass more information to the target is determined globally by the source components. In dynamic routing for aggregation, the route is iteratively updated, the more iteration have passed, the more global information are considered. While in eigen-centrality attention, the weight of each component is considered with a ranking procedure via computing the eigen-centrality.

 \section{Conclusion}

  In this work, we have proposed a new approach of aggregation, which we call eigen-centrality self-attention. Instead of computing attention score for each word with a sole dependency on the word itself as done in the vanilla self-attention, eigen-centrality self-attention computes a score for each word and taking the whole sequence into consideration. It achieved that by build a fully-connected word graph, in which the connected relationship is described by an adjacency matrix.  We can compute the eigen-centrality measure of each word in the graph which indicates how important it is in the graph. Thus it will output attention score for each word with global information considered. Since we apply power method to obtain the eigenvector, it is non-trivial to compute the gradient for the power method operation, we can compute the gradient with auto-differentiation tools, but this brings other problems such as large memory consumption and considerable computation overhead. Thus we introduced two ways of compute gradient with respect to the adjacency matrix. The first is to run extra iterations just for the step without storing intermediate states in convergence phase, this reduce considerable amount of memory consumption and reduce some computation consumption. The second way it to compute gradient after convergence phase analytically without storing any intermediate states (no extra gradient iterations needed), this reduces memory and computation consumption even more. Also note that since we are taking an iterative approach for the eigenvector it is expectable that the approach is time demanding, especially in the case where the sentence is longer. When experimented on Yelp2014 we observed 5 times more time are required than that of attentive pooling.

  When applying eigen-centrality self-attention to encode a sentence, we have the byproduct --- the word graph is also very interesting, in the future we expect to drive our aggregation approach with more supervised data such as large parallel translation datasets and devise some unsupervised task if possible, and expect to see some more interesting graph patterns in natural language.

\section*{Acknowledgement}
    This work is supported by the National Natural Science Foundation of China under Grant Nos.~61751201 and 61672162, the Shanghai Municipal Science and Technology Major Project under Grant No.2018SHZDZX01 and ZJLab.


\appendix
\section{Appendix}
  \subsection{Gradient of Power Method}
   \label{ap:grad}
   Given an adjacency matrix $\bA$ with all its' components to be strictly positive, We can obtain a good approximation of dominant eigenvector $\balpha$ and its' corresponding eigenvalue $\lambda$ via power method.
   Now we need to compute the partial derivative of $\balpha$ with respect to the eigenvector $\bA$, the power method can be formulated as follows:

   \begin{align}
      \balpha^{t} &= \frac{\bA^{t-1}\balpha^{t-1}}{\norm{\bA^{t-1}\balpha^{t-1}}_2}
   \end{align}
   where $\balpha^t$ is the $t$th step of eigenvector, and $\bA^{t-1}$ is the $(t-1)$th adjacency matrix, the value between $\bA^{t}$ and $\bA^{t-1}$ is the same, since the power method is an iterative approach, we need to consider the gradient for each step and summarize them in the end. Also note that when converged and with $t$ large enough, $\balpha^t \approx \balpha^{t-1}$ and $\norm{\bA^{t-1}\balpha^{t-1}}_2 \approx \norm{\bA\balpha}_2 \approx \lambda$.

   Assuming $t$ is large enough and iteration long converged, first inspect the partial derivative of $\alpha^{t}_p$ with respect to $A^{t-1}_{qr}$ to obtain the jacobian of $\balpha^{t}$ wrt. $\bA^{t-1}$ . For simplicity, we denote $\bA^{t-1}$ as $\bM$, and since we need only the value of $\balpha^{t}$ and $\balpha^{t-1}$ we denote then as $\balpha$:

   \begin{align}
     \alpha_p &= \frac{\sum_i{M_{pi} \alpha_i}}{\norm{\bM\balpha}_2}
   \end{align}

   \begin{align}
     \frac{\partial \alpha_p}{\partial M_{qr}} &= \frac{\frac{\partial \sum_i{M_{pi}\alpha_i}}{\partial M_{qr}}\norm{\bM\balpha}_2 - \frac{\partial \norm{\bM\balpha}_2}{\partial \bM_{qr}}\sum_i{M_{pi}\alpha_i}}{(\norm{\bM\balpha}_2)^2}
   \end{align}
   Replace $\norm{\bM\balpha}_2$ with $\lambda$, and $\sum_i{M_{pi}\alpha_i}$ with $\lambda \alpha_p$:

   \begin{align}
     \frac{\partial \alpha_p}{\partial M_{qr}} &= \frac{\lambda\frac{\partial \sum_i{M_{pi}\alpha_i}}{\partial M_{qr}} - \lambda \alpha_p\frac{\partial \norm{\bM\balpha}_2}{\partial M_{qr}}}{\lambda^2},  \label{eq:partial}
   \end{align}
   \begin{align}
     \frac{\partial \sum_i{M_{pi}\alpha_i}}{\partial M_{qr}} &= \mathbbm{1}_{p=q} \alpha_r, \label{eq:partial1} \\
     \frac{\partial \norm{\bM\balpha}_2}{\partial M_{qr}} &= \frac{1}{2\norm{\bM\balpha}_2} \frac{ \partial \sum_i{(\sum_j{M_{ij}\alpha_j})^2}}{\partial M_{qr}}, \\
     &=\frac{1}{2\lambda}\cdot 2(\sum_j{M_{qj} \alpha_j})\alpha_r, \\
     &=\alpha_q \alpha_r. \label{eq:partial2}
   \end{align}
   Plug Eq. \ref{eq:partial1} and Eq. \ref{eq:partial2} into Eq. \ref{eq:partial}:

   \begin{align}
    \frac{\partial \alpha_p}{\partial M_{qr}} &=   \frac{\mathbbm{1}_{p=q} \alpha_r - \alpha_p \alpha_q \alpha_r}{\lambda}
   \end{align}
   We can gather the partial derivative $\frac{\partial \alpha_p}{\partial M_{qr}}$ to a jacobian matrix $\bJ_{\bA} \in \mathbbm{R}^{n \times n \times n}$.

   We then inspect the partial derivative of $\alpha^{t}_p$ with respect to $\alpha^{t-1}_q$, Also we can denote $\bA^{t-1}$ as $\bM$:

   \begin{align}
     \alpha^{t}_p &= \frac{\sum_i{M_{pi} v^{t-1}_i}}{\norm{\bM\balpha}_2}
   \end{align}

   \begin{align}
     \frac{\partial \alpha^{t}_p}{\partial \alpha^{t-1}_q} &= \frac{\frac{\partial \sum_i{M_{pi}v^{t-1}_i}}{\partial v^{t-1}_{q}}\norm{\bM\balpha^{t-1}}_2 - \frac{\partial \norm{\bM\balpha^{t-1}}_2}{\partial v^{t-1}_{q}}\sum_i{M_{pi}v^{t-1}_i}}{(\norm{\bM\balpha^{t-1}}_2)^2}
   \end{align}

   Replace $\norm{\bM\balpha^{t-1}}_2$ with $\lambda$, and $\sum_i{M_{pi}v^{t-1}_i}$ with $\lambda \alpha^{t-1}_p$:
   \begin{align}
     \frac{\partial \alpha^{t}_p}{\partial \alpha^{t-1}_q} &= \frac{\lambda\frac{\partial \sum_i{M_{pi}v^{t-1}_i}}{\partial \alpha^{t-1}_q} - \lambda \alpha^{t-1}_p\frac{\partial \norm{\bM\balpha}_2}{\partial \alpha^{t-1}_q}}{\lambda^2} \label{eq:v-partial}
   \end{align}

   \begin{align}
      \frac{\partial \sum_i{M_{pi}v^{t-1}_i}}{\partial \alpha^{t-1}_q} &= M_{pq}, \label{eq:v-partial1}
   \end{align}

   \begin{align}
    \frac{\partial \norm{\bM\balpha}_2}{\partial \alpha^{t-1}_q} &= \frac{1}{2\norm{\bM\balpha}_2} \frac{ \partial \sum_i{(\sum_j{M_{ij}\alpha_j})^2}}{\partial \alpha^{t-1}_q}, \\
    &= \frac{1}{2 \lambda} \cdot 2 \sum_i{M_{iq}\sum_j{M_{ij}v^{t-1}_j}}, \\
    &= \frac{1}{2 \lambda} \cdot 2 \lambda \sum_i{M_{iq}v^{t-1}_i}, \\
    &= (\balpha^{t-1})^\top \bM_{\cdot q} \label{eq:v-partial2}
   \end{align}

   Plug Eq. \ref{eq:v-partial1} and \ref{eq:v-partial2} into Eq. \ref{eq:v-partial}:

   \begin{align}
     \frac{\partial \alpha^{t}_p}{\partial \alpha^{t-1}_q} &= \frac{M_{pq} - \alpha^{t-1}_p (\balpha^{t-1})^\top \bM_{\cdot q}}{\lambda}
   \end{align}

   Denote $\frac{\partial \alpha^{t}_p}{\partial \alpha^{t}_q}$ as $J_{pq}$, and since $\balpha^t \approx \balpha^{t-1}$, for simplicity we denote them as $\balpha$:
   \begin{align}
     J_{pq} = \frac{M_{pq} - \alpha_p \balpha^\top \bM_{\cdot q}}{\lambda},
   \end{align}
    rearrange $J_{pq}$ to a jacobian matrix $\bJ_{\balpha} \in \mathbbm{R}^{n \times n}$:
    \begin{align}
      \bJ_{\balpha} = \frac{\bM - \balpha \balpha^\top \bM}{\lambda}
    \end{align}
  Finally we can apply chain rule to get the gradient of loss $\cL$ with respect to adjacency matrix $\bA$.
  Assuming that we have a gradient from loss $\cL$ with respect to the converged eigenvector:
  \begin{align}
    \boldsymbol\gamma &= \frac{\partial \cL}{\partial \balpha} \in \mathbbm{R}^{n}
  \end{align}

  \begin{align}
    \frac{\partial L}{\partial \bA} = \lim_{N \to \infty} \sum_{k=0}^{N}{{\boldsymbol\gamma}^{\top} (\bJ_{\balpha})^k \bJ_{\bA}} \label{eq:a-grad}
  \end{align}

  Perron-Frobenius Theorem \cite{perron1907theorie, frobenius1912matrizen} says that
  \begin{align}
    \lim_{N \to \infty} \frac{\bM^N}{\lambda^N} = \balpha \bw^{\top},
  \end{align}
  where, $\balpha$ is the right dominant eigenvector and $\bw$ is the left dominant eigenvector, $\balpha$ is normalized so that $\norm{\balpha}_2=1$ and $\bw$ is normalized so that $\bw^{\top}\balpha = 1$. Then,

  \begin{align}
    \label{eq:power-convege}
    \lim_{N \to \infty} (\bJ_{\balpha})^N &= \lim_{N \to \infty} \frac{\bM^N - \balpha \balpha^\top \bM^N}{\lambda^N},\\
    &= \mathbf{0}
  \end{align}

  \cite{hua1999new} have proven the power method converge exponentially by the order of $(\frac{\lambda_2}{\lambda_1})^N$ where $\lambda_1$ is the dominant eigenvector and $\lambda_2$ is the second eigenvector, $0 < \frac{\lambda_2}{\lambda_1} < 1$, which means that Eq. \ref{eq:power-convege} converge exponentially to $0$. thus the series in Eq. \ref{eq:a-grad} is have an upper bound. Also Eq. \ref{eq:a-grad} tells us that we can take an iterative approach to get the gradient and practically we don't meed much iteration to get a good enough approximation.

  \subsection{Analysis of Power Method Initialization}
    \label{ap:zero-grad}
    We prove that the gradient is irrelevant with respect to the randomly initialized vector, we can prove that by proving that the initial vector get a gradient of zero. Assuming we have run power method for $N$ steps, then the initial vector $\bz$ received a gradient as follows:

    \begin{align}
      \frac{\partial \cL}{\partial \bz} = {\boldsymbol\gamma}^{\top} \bJ_{\balpha}^N
    \end{align}
    by using mathematical induction, we can easily prove that
    \begin{align}
      \bJ_{\balpha}^N = \frac{\bM^N - \balpha \balpha^\top \bM^N}{\lambda^N}
    \end{align}

    \begin{align}
      \frac{\partial \cL}{\partial \bz} = ({\boldsymbol\gamma}^{\top} - {\boldsymbol\gamma}^{\top}\balpha \balpha^\top)\frac{\bM^N}{\lambda^N}
    \end{align}

    \begin{align}
      \lim_{N \to \infty}\frac{\partial \cL}{\partial \bz} = ({\boldsymbol\gamma}^{\top} - {\boldsymbol\gamma}^{\top}\balpha \balpha^\top) \lim_{N \to \infty} \frac{\bM^N}{\lambda^N}
    \end{align}

    Perron-Frobenius Theorem \cite{perron1907theorie, frobenius1912matrizen} says that
    \begin{align}
      \lim_{N \to \infty} \frac{\bM^N}{\lambda^N} = \balpha \bw^{\top},
    \end{align}
    where, $\balpha$ is the right dominant eigenvector and $\bw$ is the left dominant eigenvector, $\balpha$ is normalized so that $\norm{\balpha}_2=1$ and $\bw$ is normalized so that $\bw^{\top}\balpha = 1$. Then we have:
    \begin{align}
      \frac{\partial \cL}{\partial \bz} &= ({\boldsymbol\gamma}^{\top} - {\boldsymbol\gamma}^{\top}\balpha \balpha^\top) \balpha\bw^{\top}, \\
      &= {\boldsymbol\gamma}^{\top} \balpha\bw^{\top} - {\boldsymbol\gamma}^{\top}\balpha \balpha^\top \balpha\bw^{\top} = \mathbf{0}
    \end{align}

    This suggests that if we were to compute the gradient via auto-differentiation tools such as tensorflow, when the convergence is slow, we can stop gradient before converge, and run extra steps just to get the gradients. This could save a substantial amount of memory. When the convergence is quick, the extra gradient steps could also help to accurately approximate the gradient.

     The better way of approximating the gradient is shown in Eq. \ref{eq:a-grad}, via this iterative approach, no intermediate state are required to be stored. this makes the memory complexity to be $O(1)$, and extra gradient steps are not needed.

  \subsection{Avoid a Perpendicular Initialization}
    \label{ap:analy2}
    \cite{cavender1993use} and \cite{hogben1987elementary} says that even when the adjacency matrix is defective, with it's dominant eigenvalue to have algebraic and geometric multiplicities both to be one, the power method would still converge. In this section, we suppose that the adjacency matrix $\bA$ is a non-defective matrix. And justify the choice of initialize the vector $\bz$ to have all positive components.

    Assuming that strictly positive adjacency matrix $A$ have a full set of basis eigenvectors $\bu_1, \bu_2, \cdots, \bu_n$ and corresponding eigenvalues $\lambda_1, \lambda_2, \cdots, \lambda_n$ with $\lambda_1$ to be the dominant eigenvalue such that $\lambda_1 > \abs{\lambda_2} > \cdots > \abs{\lambda_n}$. There exists real numbers $c_1, c_2, \cdots, c_n$ such that

    \begin{align}
      \bz = c_1\bu_1 + c_2\bu_2 + \cdots + c_n\bu_n
    \end{align}

    Then,
    \begin{align}
      \bA^N\bz &= c_1\bA^N\bu_1 + c_2\bA^N\bu_2 + \cdots + c_n\bA^N\bu_n, \\
               &= c_1\lambda_1^N\bu_1 + c_2\lambda_2^N\bu_2 + \cdots + c_n\lambda_n^N\bu_n
    \end{align}

    \begin{align}
      \frac{\bA^N\bz}{\lambda_1^N} &= c_1\bu_1 + c_2(\frac{\lambda_2}{\lambda_1})^N\bu_2 + \cdots + c_n(\frac{\lambda_n}{\lambda_1})^N\bu_n
    \end{align}
    Since $\forall i \quad \frac{\lambda_i}{\lambda_1} < 1$ then $\lim_{N \to \infty}(\frac{\lambda_i}{\lambda_1})^N = 0$,
    \begin{align}
      \label{eq:power-converge}
      \lim_{N \to \infty} \frac{\bA^N\bz}{\lambda_1^N} = c_1\bu_1
    \end{align}

    Eq. \ref{eq:power-converge} shows that when $c_1$ is not zero, the power method would converge to the dominant eigenvector, when $c_1$ is zero, the power method is unable to converge to the dominant eigenvector. But luckily in this paper, the dominant eigenvector is known to have all components to be real positive. This suggests that when we initialize a vector $\bz$ to have all components to be positive, we can guarantee that $c_1$ to be non-zero and possibly converge faster.


\end{document}


\ArticleType{Supplementary File}

\title{Title}{Title for citation}

\author[1]{Aaa AUTHOR}{}
\author[1,2]{Bbb AUTHOR}{{bauthor@xxx.com}}
\author[2]{Ccc AUTHOR}{}
\author[3]{Ddd AUTHOR}{}

\AuthorMark{Author A}

\AuthorCitation{Author A, Author B, Author C, et al}


\address[1]{Affiliation, City {\rm 000000}, Country}
\address[2]{Affiliation, City {\rm 000000}, Country}
\address[3]{Affiliation, City {\rm 000000}, Country}

\maketitle


\begin{appendix}

\section{Importance}
Please use this sample as a guide for preparing your letter. Please read all of the following manuscript preparation instructions carefully and in their entirety. The manuscript must be in good scientific American English; this is the author's responsibility. All files will be submitted through our online electronic submission system at \url{https://mc03.manuscriptcentral.com/scis}.

\section{More information}
The examples at the bottom of the .tex file can help you when preparing your manuscript. We are appreciate your effort to follow our style~\cite{1,2}.

\end{appendix}
